\documentclass{bmvc2k}

\usepackage{pifont}
\usepackage{adjustbox}
\usepackage{wrapfig}

\newcommand{\ourmodel}{{VisualSplit}}  % \fontfamily{ppl}\selectfont

\newcommand{\NAME}{VisualSplit}

\definecolor{linkblue}{rgb}{0.21,0.49,0.74}
\newcommand{\linkblue}[1]{\textcolor{linkblue}{#1}}

\usepackage{xcolor}         % colours
\usepackage{graphicx}
\usepackage{booktabs}

\definecolor{cblue}{rgb}{0.21,0.49,0.74}
\definecolor{cvprblue}{rgb}{0.21,0.49,0.74}
\usepackage{multirow,multicol,amssymb,makecell,diagbox,colortbl,amsmath,amsthm,pifont}
\usepackage{times,soul,url,algorithm,algorithmic}

\usepackage{graphicx}
\usepackage{booktabs}
\usepackage{amssymb}   % http://ctan.org/pkg/amssymb
\usepackage{pifont}    % http://ctan.org/pkg/pifont
\usepackage{wrapfig}
\usepackage{makecell}
\usepackage{amsmath}
\usepackage{bm}
\usepackage{multirow}

\usepackage{colortbl}
\usepackage{xcolor}  % [dvipsnames]

\usepackage{makecell}
\usepackage{amsmath}
\usepackage{bm}
\usepackage{multirow}
\usepackage{scalerel}  % Scale in Height

\usepackage{caption}
\usepackage{subcaption}
\usepackage{enumitem}

\usepackage{epigraph}   % 简洁的引言/题签
\usepackage{csquotes}   % 优雅的引号（可选）

\usepackage[accsupp]{axessibility}

\title{Exploring Image Representation with Decoupled Classical Visual Descriptors}

\addauthor{Chenyuan Qu}{cxq134@student.bham.ac.uk}{1,3}
\addauthor{Hao Chen}{hc666@cam.ac.uk}{1,2}
\addauthor{Jianbo Jiao}{j.jiao@bham.ac.uk}{1}

\addinstitution{
 The MIx Group\\
 University of Birmingham,
 UK
}
\addinstitution{
 University of Cambridge\\
 Cambridge, UK
}
\addinstitution{
 Allsee Technologies Ltd\\
 Birmingham, UK
}
\runninghead{Chenyuan Qu, Hao Chen, Jianbo Jiao}{Exploring Image Representation}

\def\eg{\emph{e.g}\bmvaOneDot}

\def\etal{\emph{et al}\bmvaOneDot}

\begin{document}
\maketitle

\definecolor{pinksalmon}{rgb}{0.8, 0.5, 0.5}

\begin{abstract}
% TODO

Exploring and understanding efficient image representations is a long-standing challenge in computer vision.  
While deep learning has achieved remarkable progress across image understanding tasks, its internal representations are often opaque, making it difficult to interpret how visual information is processed. In contrast, classical visual descriptors (\eg edge, colour, and intensity distribution) have long been fundamental to image analysis and remain intuitively understandable to humans. 
Motivated by this gap, we ask a central question: \textit{Can modern learning benefit from these classical cues?} In this paper, we answer it with \ourmodel, a framework that explicitly decomposes images into decoupled classical descriptors, treating each as an independent but complementary component of visual knowledge. Through a reconstruction-driven pre-training scheme, \ourmodel~learns to capture the essence of each visual descriptor while preserving their interpretability. 
By explicitly decomposing visual attributes, our method inherently facilitates effective attribute control in various advanced visual tasks, including image generation and editing, extending beyond conventional classification and segmentation, suggesting the effectiveness of this new learning approach for visual understanding. 
Project page: \url{https://chenyuanqu.com/VisualSplit/}.

\epigraph{\makebox[0.9\textwidth][c]{\hspace{-12mm}``To see a world in a grain of sand.''}}{\textit{--- William Blake}}

% \epigraph{\emph{\enquote{To see a World in a Grain of Sand.}}}
         % {\textit{— William Blake}}
% Decomposing visual components 

\end{abstract}

% \section{TODO list}
% 1. Figure and Table make clean,  by 14 May @ Hao

% 2. rebuttal points

% 3. information entropy table @ Chenyuan

% 4. Related work from rebuttal @ Waiting

\section{Introduction}
\label{sec:intro}

Computer vision, and more recently deep learning, has continuously pursued effective  {\textit{decomposition}} of visual information, 
to extract semantically meaningful components directly from images, enabling a deeper and clearer understanding of their content. In classical computer vision, feature extraction relied heavily on predefined methods such as dimensionality reduction~\cite{macqueen1967some}, low-level feature descriptors~\cite{kanopoulos1988design}, and handcrafted feature extractors~\cite{lowe2004distinctive, dalal2005histograms}.
While these traditional methods offer strong deterministic properties, they are limited in adaptability and scalability, as they are tailored to specific tasks without the ability to generalise across diverse scenarios.

% \begin{figure*}[!t]

% %  左下右上
% \newcommand\Ma{\includegraphics[width=0.23 \textwidth]}  

\begin{figure}[t]
    \centering
    \vspace{-0.05cm}
    \begin{subfigure}[b]{0.23\textwidth}
        \includegraphics[width=\textwidth]{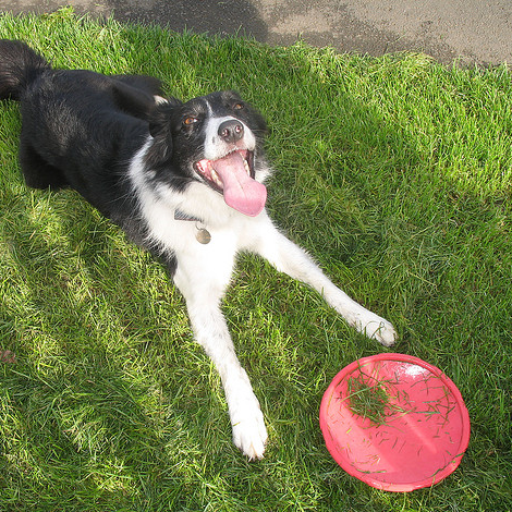}
        \caption{Original Image}
        \label{fig:raw image}
    \end{subfigure}
    ~ % 添加这个符号来使子图并排显示，而不是换行
    \begin{subfigure}[b]{0.23\textwidth}
        \includegraphics[width=\textwidth]{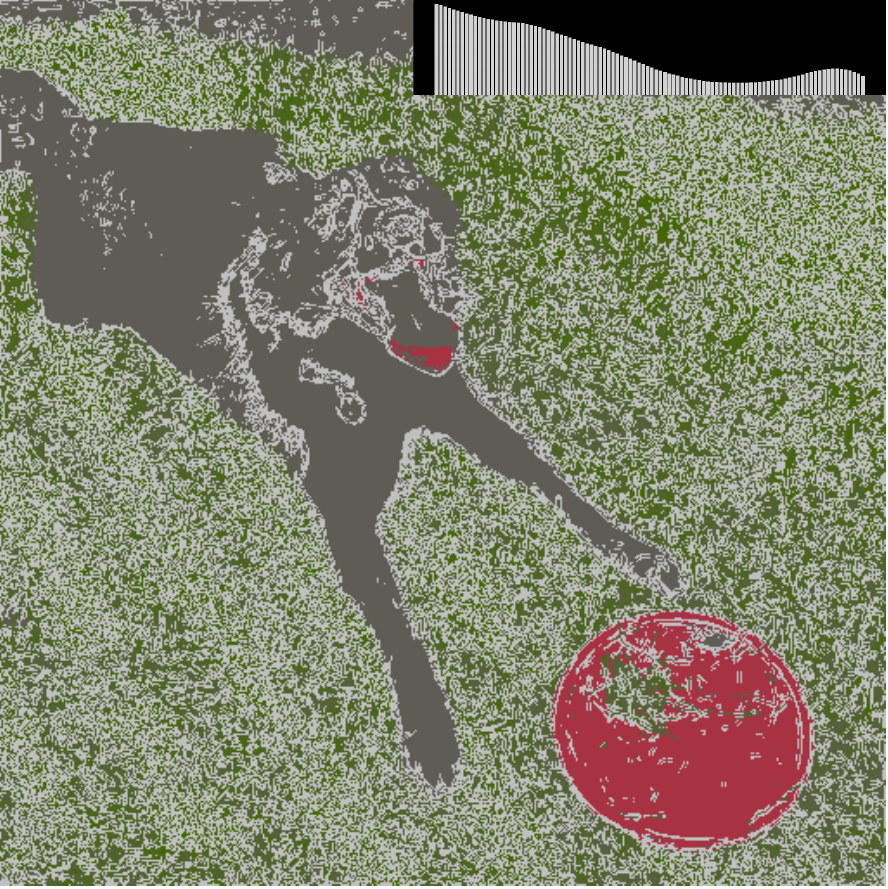}
        \caption{Combined Features}
        \label{fig:extracted feature combination}
    \end{subfigure}
    ~ % 同上，继续保持子图并排  %  左下右上
    \begin{subfigure}[b]{0.23\textwidth}
        \includegraphics[width=\textwidth, trim=0 0 33 0,clip]{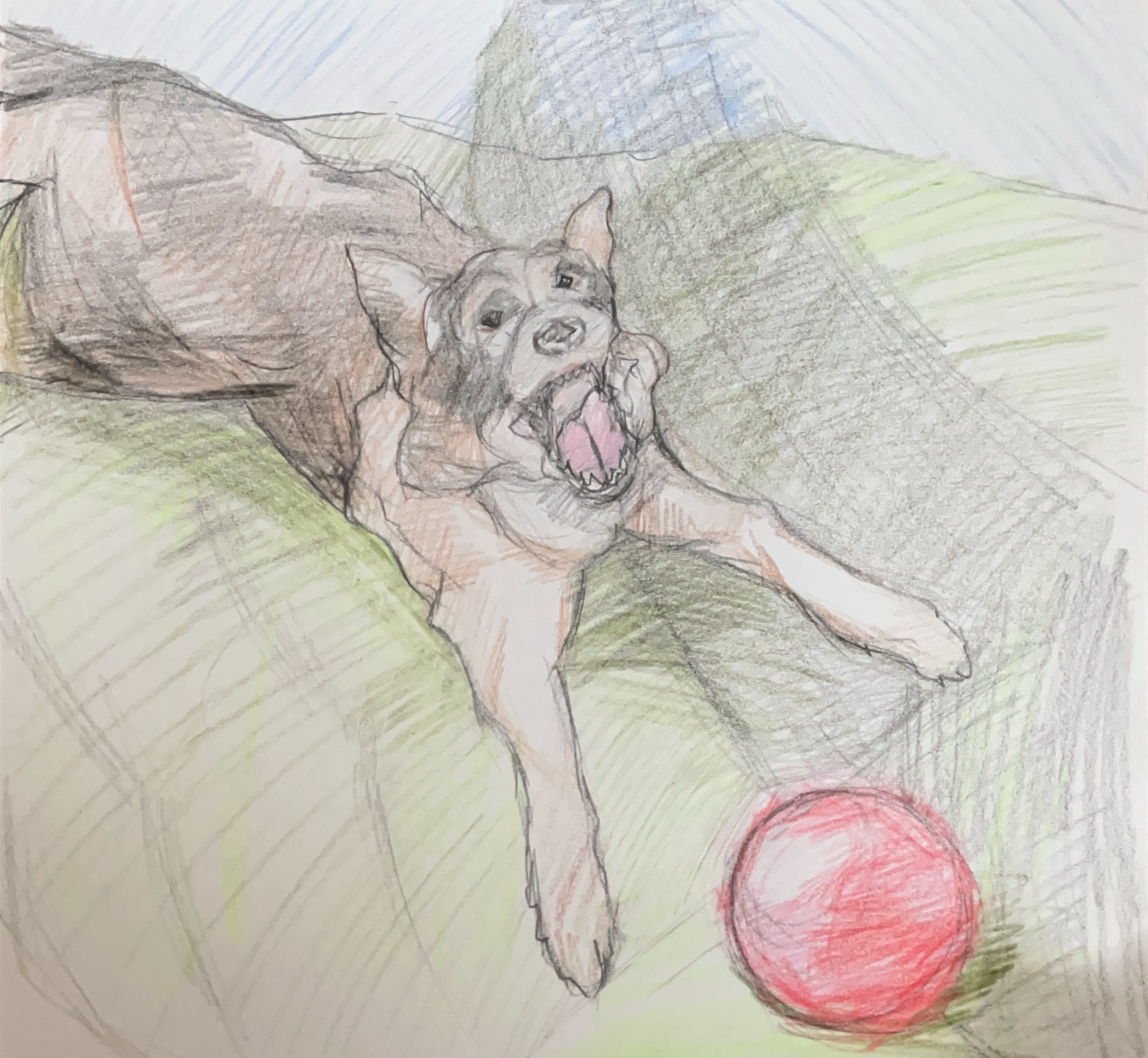}
        \caption{Human Depicted\footnotemark}
        \label{fig:human artist}
    \end{subfigure}
    ~ % 继续保持子图并排
    \begin{subfigure}[b]{0.23\textwidth}
        \includegraphics[width=\textwidth]{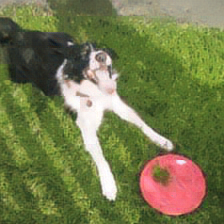}
        \caption{\NAME~(Ours)}
        \label{fig:model generationed}
    \end{subfigure}
    \vspace{0.25cm}
    \caption{\textbf{Qualitative illustration of the key idea.} \textbf{(a)} Original image. \textbf{(b)} Visualisation of the threefold visual descriptors (grey-level histogram displayed in the top-right). \textbf{(c)} Image drawn by a human artist using (by looking at) only the combined descriptor set in (b). 
    \textbf{(d)} Recovery from the proposed \NAME, only using the same combined descriptor set.}
    \label{fig:illustration}
    \vspace{-0.4cm}
\end{figure}
 \footnotetext{We thank Jie Dong for the illustration in panel (c); used with permission.}

% as its sole input

% triplet,  from ({b})

In contrast, contemporary deep learning strategies approach visual decomposition through learning-based paradigms such as supervised learning~\cite{weideep, li2021image, domaingame,Chen_2023_ICCV} and self-supervised pretext learning~\cite{zhang2017split}. For instance, RetinexNet~\cite{weideep} decomposes images into reflectance and illumination components, enabling controlled lighting adjustments while preserving colour.
Similarly, HiSD~\cite{li2021image} effectively isolates essential human facial attributes, facilitating fine-grained control over expressions and facial feature modifications. More broadly, the split-brain autoencoder \cite{zhang2017split}  demonstrates image decomposition by partitioning visual inputs according to the LAB colour channels, enhancing feature extraction and understanding.

% \textbf{``Whether the interpretability of classic visual descriptors as a foundation can improve the learning of robust and disentangled  representations?"} 

\smallskip\noindent\textbf{Main idea:} Despite extensive studies on classical visual descriptors from traditional computer vision with well-established deterministic properties, their potential in the context of deep learning models has been largely overlooked. In this paper, we are interested in the question:  \textit{``Can classical visual descriptors be used to decompose visual information in learning-based frameworks?"}.   
 From an artistic perspective, images are characterised by stylistic descriptors such as \ul{\textit{line, colour, and value}} \cite{inbook}, which closely align with the key components for computational visual understanding and analysis \cite{boyle1988computer}. Inspired by these parallels, we seek to leverage 
conventional computer vision algorithms to extract visual descriptors—specifically \ul{\textit{edge, colour segmentation map and grey-level histogram}} (Section~\ref{sec:methods}). 

The choice of descriptors is motivated by our aim to intuitively separate complementary and meaningfully isolated components within the image.
We select three canonical descriptors: edges, which capture local geometric structure; colour segmentation, which encodes region-wise chromatic structure; and grey-level histograms, which summarise global photometric statistics. These descriptors also encode varying degrees of spatial specificity: edges preserve fine spatial detail; histograms capture non-spatial intensity distributions; and colour segmentation bridges the two by grouping pixels into coherent chromatic regions. We highlight that a comprehensive exploration of all possible descriptors lies beyond the scope of this study; we view this classical, representative set as a principled starting point and encourage follow-up work to expand it. Notably, the combined descriptor depiction offers only partial information about the image (\figureautorefname~\ref{fig:extracted feature combination}).

Although these descriptors are highly abstract and compressed compared to the original image, humans can still figure out the main underlying semantic content. To this end, we present a case study in which an artist was asked to depict an image given such ``abstract information''. Interestingly, the artist was able to reconstruct an image (\figureautorefname~\ref{fig:human artist}) very similar to the original one (\figureautorefname~\ref{fig:raw image}). We attribute this ability to the human capability for understanding incomplete visual concepts. Motivated by this, we seek to explore whether such an ability can be modelled by a network. To achieve that, we introduce \ourmodel, a mask-free framework that follows such visual understanding process by learning to integrate only these classical descriptors to reconstruct the underlying image (\figureautorefname~\ref{fig:model generationed}), which achieves high representation quality while bringing extra controllability (details in Section \ref{sec:discussion}).

% This example illustrates the remarkable capacity of human cognition to infer missing details by leveraging prior knowledge and a holistic understanding of visual concepts. 

\smallskip\noindent\textbf{Remark:} Although \ourmodel~is implemented in a self-supervised manner, the primary focus of this work is on the study of decoupled visual representations rather than conventional evaluation pipelines in self-supervised representation learning literature. Our goal is not to surpass existing representation learning methods on those benchmarks, but rather aims to answer the question posed above. 
% In summary, our contributions can be summarised as follows:
In summary, our key contributions are as follows:
% To answer why three component
\begin{itemize}[topsep=0pt, itemsep=1pt, parsep=0pt]  % , left=2pt

    \item Our \NAME~framework enhances image decoupling quality, achieving consistently better performance across high-level (Sec.~\ref{subsec:classification}) and low-level vision tasks (Sec.~\ref{subsec:transfer learning}), showing the robustness and generalisability of decomposed visual descriptors.

    \item \NAME~implicitly facilitates the decomposition of learned attributes through decoupled visual inputs, enabling precise, independent, and intuitive manipulation of image attributes such as geometry, colour, and illumination, as validated in image generation and editing tasks (Sections~\ref{subsec: generation} and \ref{subsec: editing}).

\end{itemize}

\section{Related Work}
\label{sec:related work}
\vspace{-0.2cm}

\smallskip \noindent\textbf{Visual descriptors.}
In the field of computer vision, visual descriptors are widely used to represent image content. In the classic computer vision area, traditional visual descriptors are manually designed to capture specific image properties, including edge maps~\cite{sobel19683x3}~\cite{canny1986computational}, colour histograms~\cite{szeliski2022computer}, Scale-Invariant Feature Transform (SIFT)~\cite{lowe2004distinctive}, and Histogram of Oriented Gradients (HOG)~\cite{dalal2005histograms}. Despite their usefulness and deterministic nature, traditional descriptors suffer from rigidity, sensitivity to variations such as lighting and perspective changes, and limitations inherent in manual feature engineering. 
In recent years, deep learning approaches have relied mainly on learning-based methods, particularly convolutional neural networks (CNNs) and Vision Transformers (ViTs). These approaches automatically learn rich feature representations from extensive datasets, significantly improving robustness and descriptive power.~\cite{schonberger2017comparative}~\cite{bursuc2015kernel}~\cite{dong2015domain}~\cite{tian2017deepcluster} However, learning-based methods are not easily interpretable and often mixing multiple attributes and patterns.

% TODO
% Therefore, we aim to explore the potential to leverage the well-established deterministic of traditional visual descriptors and use them in learning-based methods to improve performance and robustness, which is largely neglected in the current studies.

\smallskip \noindent\textbf{Visual decoupling and disentanglement.}
\label{subsec:Feature Disentanglement and Representation Learning}
Decoupling visual features focuses on eliminating dependencies between components, whereas disentanglement emphasises the separation of distinct underlying factors that correspond to meaningful and independent concepts.  
These principles are of great importance in various applications such as visual generation~\cite{kingma2013auto, yang2021causalvae}, model generalisation~\cite{BayeSeg,chen2024domain}, and model explainability~\cite{yin2019semantics,zhang2017split}.   
Given their conceptual overlap, we discuss both lines of research in this discussion. 
The Split-Brain model~\cite{zhang2017split} decouples representations by independently processing the LAB channels, enabling the isolation of distinct channel contributions to the overall task.  
Li \etal~\cite{li2020improving} propose decoupling an image into low-frequency and high-frequency elements, which correspond to body features and edge features, respectively, with a stronger focus on semantic segmentation tasks. 
Yang \etal~\cite{yang2021causalvae} develop a causal disentanglement approach to align latent factors with the semantics of interest. 
BayeSeg~\cite{BayeSeg} disentangles domain-invariant features to enable generalisation to unseen data.

\section{Methodology}
\label{sec:methods}

% ({\color{red} Note:} The $X$ in (a) is actually $I$, will be fix in next figure updates.)
% L D R U

\begin{figure*}[t]
    \centering
    % \includegraphics[width=\textwidth, trim=23 455 423 3,
    % clip]{figures/eccv2024.pdf}
    \includegraphics[width=\textwidth]{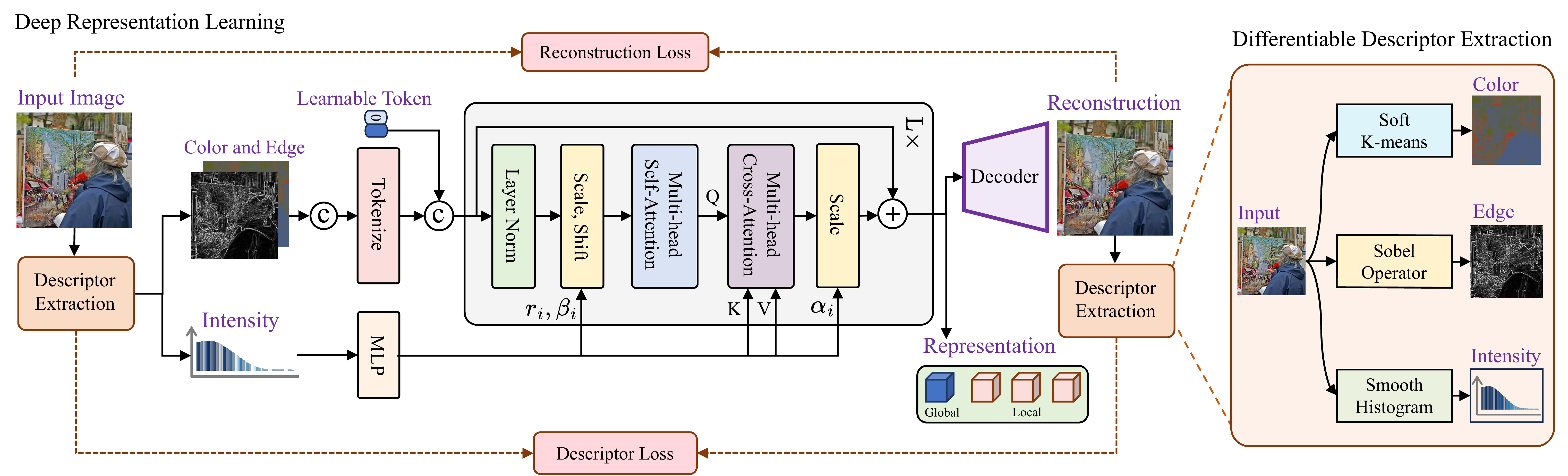}
    \vspace{1mm}
    \caption{\textbf{\NAME~framework.} 
  Initially, the RGB image is processed through the traditional \textit{Descriptor Extraction} module to obtain segmented colour, edge, and intensity histograms. The colour and edge data will be patchtified and fed into the transformer, whereas the intensity information will serve as a condition for Cross-attention and to learn scale and shift parameters ($r, \beta$) for AdaLN-Zero.
  % Used to be AdaLM
   The loss is calculated on an image level as well as on the traditional descriptor level. 
   It is worth mentioning that the acquired model can output both global and local representations, which can be applied to downstream visual tasks. 
   }
    % into the LAB domain to derive the AB and L components. During the encoding phase, the AB and L components are transformed into 4-bit AB, intensity histograms, and edges, each processed through separate Swin encoders to extract colour, illumination, and geometry features. These distinct features are consolidated in a Swin-fusion unit to create a fused feature. In the reconstruction phase, the fused feature undergoes processing through a CNN to map back to the LAB image, allowing for the extraction of four different types of outputs to calculate the loss functions. 
    %  are learned to enhance understanding for downstream visual tasks.
    \vspace{-0.4cm}
    \label{fig:arch}
\end{figure*}

As illustrated in \figureautorefname~\ref{fig:arch}, the proposed \ourmodel~approach learns decoupled representations by reconstructing the original visual signal from isolated, incomplete visual descriptors.   \NAME~comprises  three key components: descriptor extraction, a multi-modal encoder, and a lightweight decoder for image reconstruction.

% fragmentary, scattered, segmented 

% Descriptors

% Our architecture based on the ViTMAE~\cite{he2022masked}, and 

% which contains an encoder to map the visual inputs to the latent representation and a decoder to reconstruct from the latent representation. 

% Meanwhile, we adopt an asymmetric design like ViTMAE~\cite{he2022masked}, which treats edge map, low-resolution colour map, and grey-level histogram as input. Unlike ViTMAE, our input is from different sources and our outputs are in different domains, while inputs can also be local or global. Therefore, we designed a multi-modal encoder t o map all inputs to latent space, and followed by a simple transformer-based decoder to reconstruct the original image.

% Specifically, we extract structural information by applying edge detection to capture line elements, isolating the geometric framework of the image. For color representation, we use k-means clustering on the AB channels of the LAB color space, segmenting the image into color patches while minimizing structural details. Finally, we focus on value (intensity) by computing a histogram over the L channel, effectively abstracting away spatial information and isolating the tonal characteristics of the image. Each component provides an abstract, partial representation of the image, which naturally fits the pretext task learning framework, leveraging incomplete views to enhance feature learning. 

\noindent \textbf{Descriptor extraction.} The \ourmodel~leverages traditional computer vision algorithms to extract three deterministic and human-understandable descriptors: \textit{edge maps}, \textit{colour segmentation maps}, and \textit{grey-level histograms}, denoted as $d_{e}$, $d_{c}$ and $d_{g}$, respectively.

To extract the desired descriptors, we first transform the image $ x $ from the RGB domain into the LAB domain, denoted as $ x_{{}_\text{LAB}} $.
From the L-channel, we apply the Sobel operator~\cite{kanopoulos1988design} ($ d_{e} = \text{Sobel}(x_{{}_\text{L}}) $) to to obtain the edge map, which can be treated as edge descriptors. Simultaneously, we compute a grey-level intensity histogram with 100 bins ($ d_{g} = \text{Hist}(x_{{}_\text{L}}) $), using a smooth Gaussian kernel, providing the intensity distribution. 
For colour segmentation, we perform soft k-means clustering on A and B channel, which partitions the image into $ K $ clusters ($ d_{c} = \text{Cluster}(x_{{}_\text{AB}}) $) based on colour values.
Notably, all operators are implemented in a differentiable manner; see Supp. Sec. \linkblue{2.2}.

Existing self-supervised representation learning methods are typically built upon the principle of missing information prediction, while the common approach is explicit masking strategies. In contrast, our method employs classical visual descriptors, which inherently produce sparse and abstract representations of the original image. These descriptors can be regarded as a form of information absence, therefore, additional artificial patch masking is unnecessary. Moreover, our approach effectively preserves and reinforces globally and locally meaningful visual structures within the learned representations.

\noindent \textbf{Multi-modal encoder.} Visual descriptors describe images either locally or globally. In our case, edge and colour segmentation descriptors primarily capture localised pixel-level information, whereas the grey-level histogram captures global information. Given this distinction, it is essential to employ a multi-modal encoder capable of considering each type of descriptor simultaneously, while also mapping local and global information into latent spaces, ensuring a comprehensive representation.

Building on this observation, we construct a multi-modal encoder based on ViT~\cite{vaswani2017attention}, which operates on local patches. To accommodate the distinct roles of local and global descriptors, we treat local descriptors as patch-wise inputs for backbone learning, while incorporating global descriptors, such as histograms, to provide global conditioning.

The conditioning process consists of two key components: \underline{AdaLN-Zero}~\cite{peebles2023scalable} and a \underline{Multi-head Cross-attention Layer}, each operating within separate blocks. AdaLN-Zero learns parameters that scale and shift intermediate features derived from an MLP, which takes $d_g$ as input. Simultaneously, the multi-head cross-attention layer utilises $d_g$ as the key-value pair, allowing the model to integrate relevant information from a global perspective.  

To distinguish between global and local representations in the output, we introduce one learnable global token at the encoder input (replacing [CLS] token in ViT), allowing it to participate in all self-attention layers followed by the ViT~\cite{vaswani2017attention} design, while the remaining outputs correspond to local representations.

\noindent \textbf{Decoder.} Our model employs a more lightweight decoder specifically designed for the pretraining phase. The primary objective is to ensure that the encoder learns a more complex representation than the decoder. To achieve this, the decoder is intentionally kept much shallower than the encoder, a design choice that also accelerates the training process. We discuss it further in Supp. Sec. \linkblue{2.3}.

\noindent \textbf{Pre-training objectives.} 
The goal of our pre-training is to reconstruct the original image. Therefore, the primary objective during this phase is to minimise the discrepancy between the reconstructed image $ \hat{x}$ and the original image $ x $ in the pixel space. To achieve this, we compute the mean squared error (MSE) between the two images. Additionally, we incorporate the LPIPS loss function~\cite{zhang2018unreasonable} to assess reconstruction quality in latent space, providing a more comprehensive and perceptually aligned evaluation. 
% \begin{equation}
%     L_{\text{MSE}} = ||x_{{}_\text{RGB}} - \hat{x}_{{}_\text{RGB}}||_2^2, \quad\quad
%     L_{\text{LPIPS}} = \text{LPIPS}(x_{{}_\text{RGB}}, \hat{x}_{{}_\text{RGB}})
%     \label{eq:image_loss}
% \end{equation}

Furthermore, the encoder is required to effectively capture input visual descriptors and map them to latent representations. To ensure its capability to do so, we introduce a descriptor consistency loss. By applying the same processing operations to the reconstructed image in pixel space as to the input, we apply visual descriptors on the reconstructed image. The output of these descriptors is then compared with the original ones to compute the descriptor consistency loss, consisting of three terms: 
\begin{equation}
\begin{aligned}
L_{\text{e}} &= \|d_{e} - \hat{d}_{e}\|_1^1, \quad
L_{\text{g}} = \frac{1}{N} \sum_{i}^N \frac{(d_{g}^{\,i} - \hat{d}_{g}^{\,i})^2}{d_{g}^{\,i} + \hat{d}_{g}^{\,i} + \epsilon}, \quad
L_{\text{c}} = \|d_{c} - \hat{d}_{c}\|_1^1,
\end{aligned}
\label{eq:descriptor_losses}
\end{equation}
where $d_{e}$, $d_{g}$ and $d_{c}$ represent edges, grey-level histogram, and colour segmentation map, respectively. Ablation studies on objective functions are shown in Supp. Sec. \linkblue{4}.

\section{Experiments}
\label{sec:experiments}

% \begin{figure}
%     \centering
%     \includegraphics[width=\linewidth, keepaspectratio]{figures/experiments/pre-train.pdf}
%     \caption{\textbf{Visualisation of pre-train reconstruction task.} Top: original images, bottom: our restoration results.}
%     \label{fig:pre-trained}
% \end{figure}

\begin{figure}[t]
    \centering
    \begin{minipage}[t]{0.35\linewidth}
        \centering
        \includegraphics[width=\linewidth, keepaspectratio]{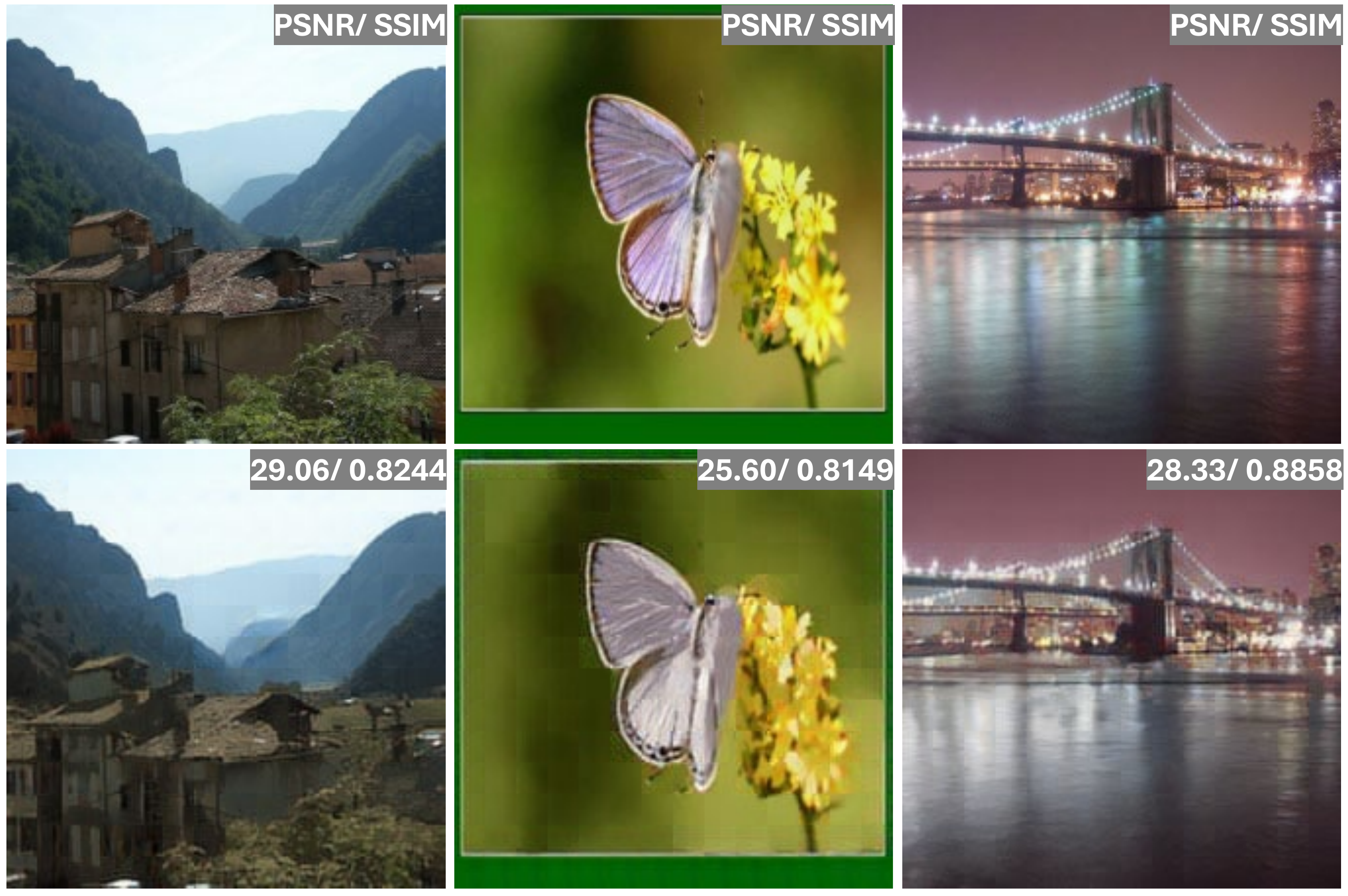}
        \vspace{1mm}
        \caption{\textbf{Visualisation of the pre-training reconstruction task.} Top: original images, bottom: our restoration results.}
        \label{fig:pre-trained}
    \end{minipage}
    \hfill
    \begin{minipage}[t]{0.64\linewidth}
        \vspace{-2.98cm}
        \centering
        \resizebox{\linewidth}{!}{
            \begin{tabular}{c@{\hspace{0.5em}}c@{\hspace{0.5em}}c@{\hspace{0.5em}}c}
                \includegraphics[width=0.5\linewidth, keepaspectratio]{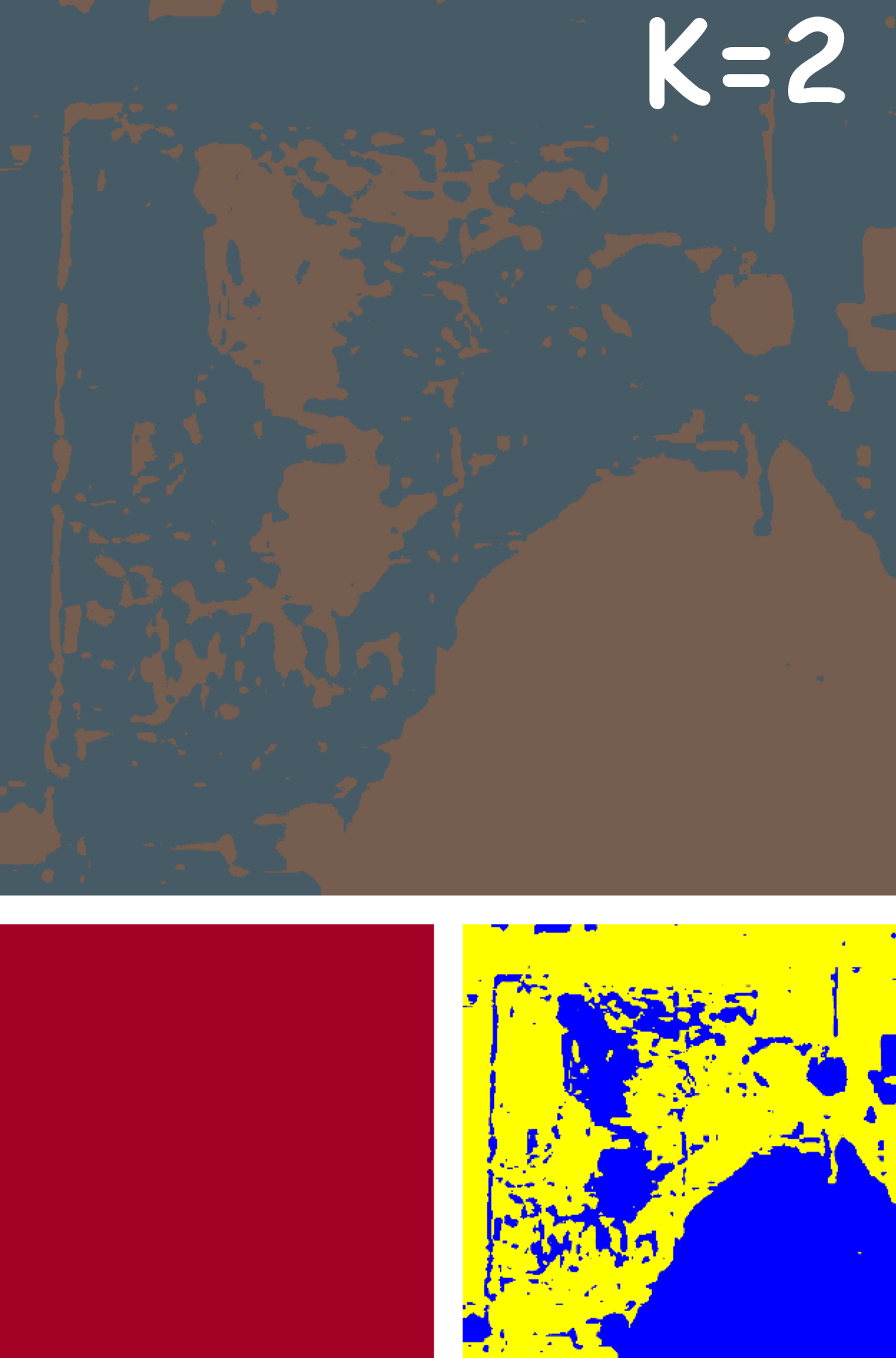} &
                \includegraphics[width=0.5\linewidth, keepaspectratio]{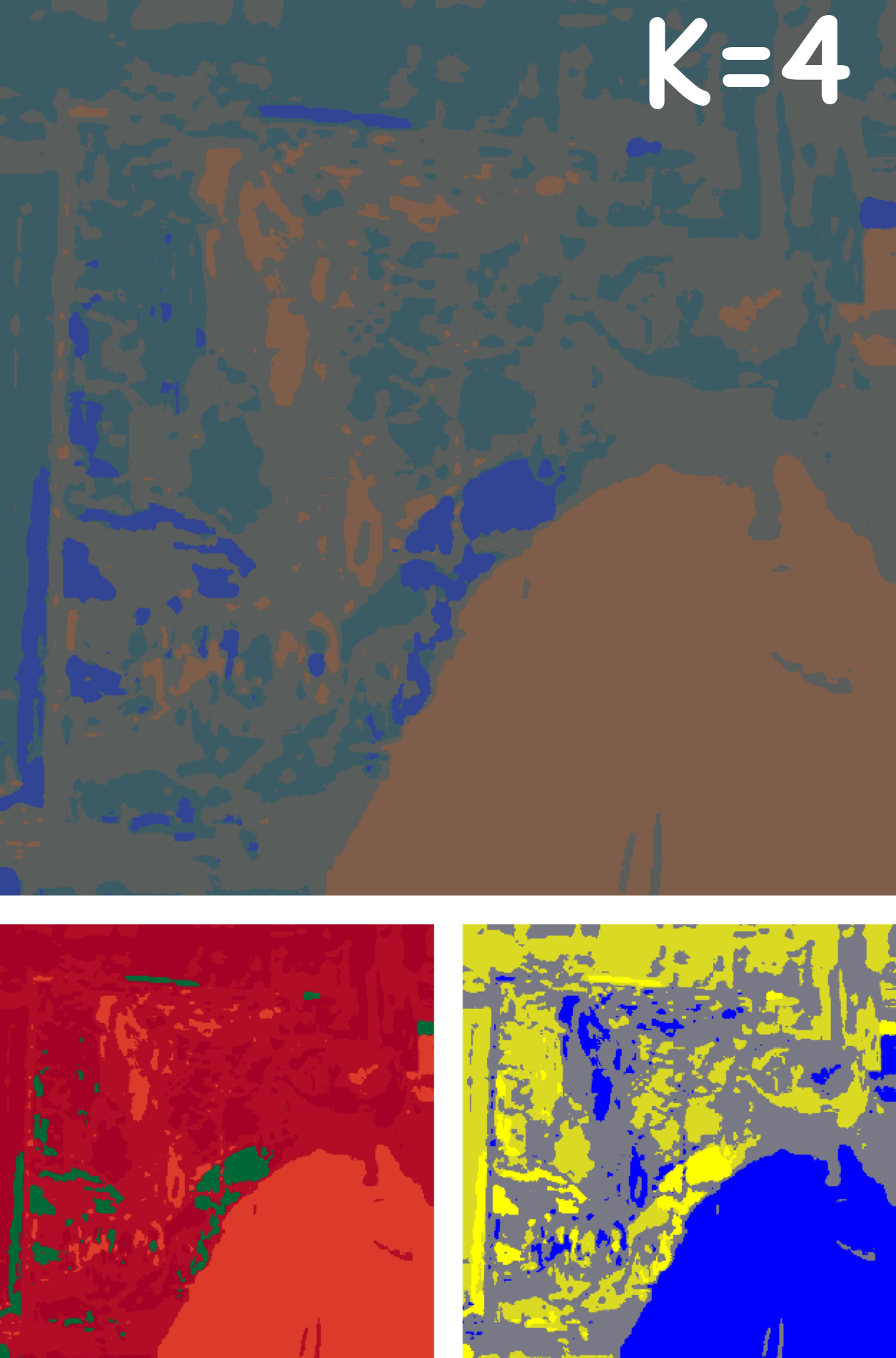} &
                \includegraphics[width=0.5\linewidth, keepaspectratio]{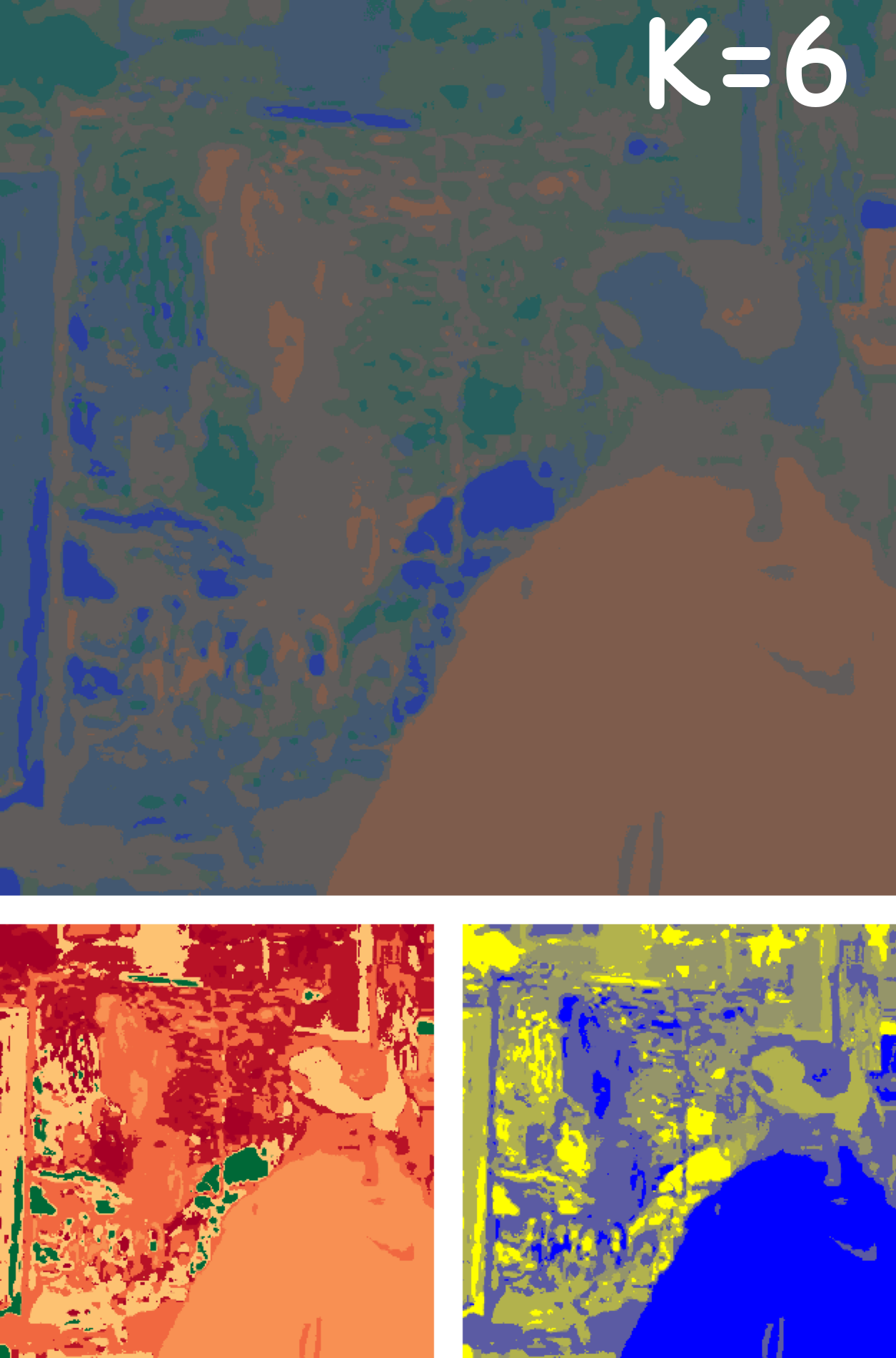} &
                \includegraphics[width=0.5\linewidth, keepaspectratio]{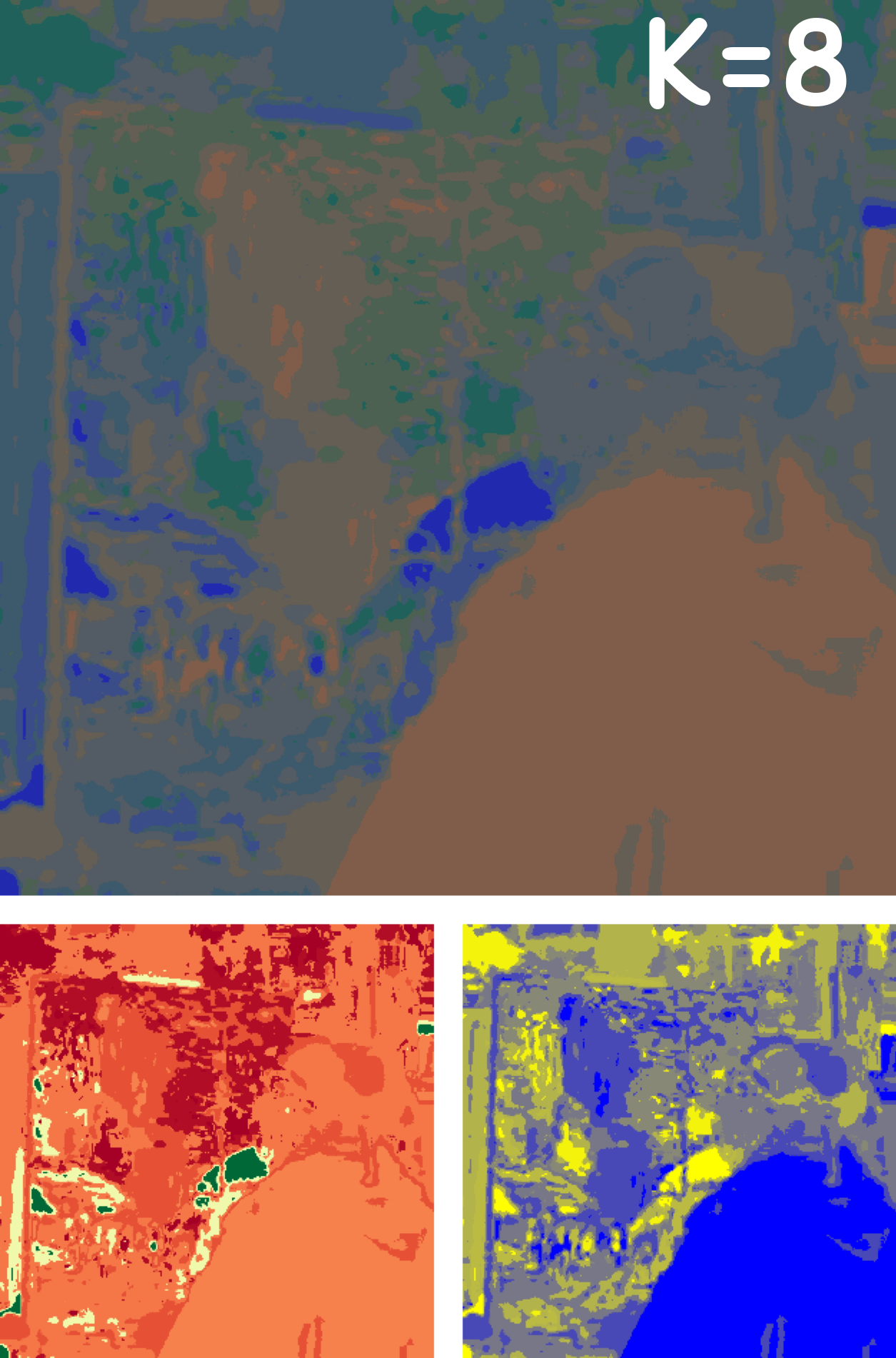} \\
                \LARGE Acc. = $60.0$ & \LARGE Acc. = $67.1$ & \LARGE Acc. = $\mathbf{74.0}$ & \LARGE Acc. = $73.4$
            \end{tabular}
        }
        \vspace{2.5mm}
        \caption{\textbf{Clustering visualisation with $K$ clusters.}  Each set shows the colour-clustered LAB image (top), alongside the isolated A-channel and B-channel (bottom left/right). Acc. indicates the classification accuracy of linear probing.}
        \label{fig:num_of_cluster_data}
    \end{minipage}
    \vspace{-0.6cm}
\end{figure}

% K-means clustering results for \(K = 2, 4, 6,\) and \(8\).

% Zoom in for a better view. 

\begin{figure}[t]
\vspace{-0.35cm}

\end{figure}

\begin{figure}[t]
    \centering
    \begin{minipage}[t]{0.36\linewidth}
        \centering
        \includegraphics[width=\linewidth, keepaspectratio, trim=0 10 0 0, clip]{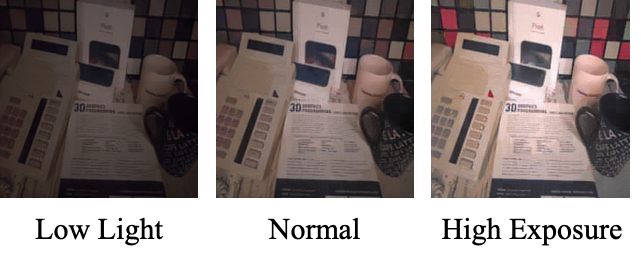}
        \vspace{0.2mm}
        \caption{\small \textbf{Reconstruction with grey-level histograms} at different brightness levels; edge and segmented colour map keep the same.}
        \label{fig:brightness}
    \end{minipage}
    \hfill
    \begin{minipage}[t]{0.62\linewidth}
        \centering
        \includegraphics[width=\linewidth]{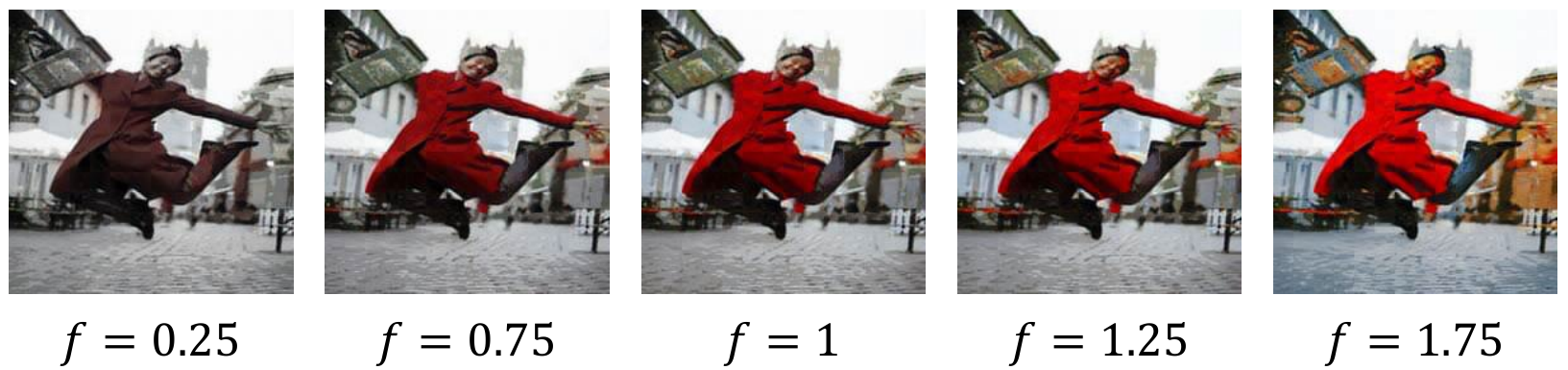}
        \vspace{0.2mm}
        \caption{\small \textbf{Reconstruction with different colour segmentation map inputs} extracted from the image after different colour balance processing, where $f$ is the enhancement factor; edge and grey-level histograms keep the same.}
        \label{fig:colour}
    \end{minipage}
    \vspace{-3mm}
\end{figure}

% In this section, we first explain the implementation details in Section \ref{subsec:implementation details}. Section \ref{subsec:pre-training} presents the pre-trained task performance. And then we assess the quality of these representations through their application in several key computer vision tasks. Specifically, classification performance i
We present classification results in Section \ref{subsec:classification}, and transfer-learning capabilities in Section \ref{subsec:transfer learning}. In Section \ref{subsec: generation}, we demonstrate how representations extracted by our model guide pre-trained generative models and Section \ref{subsec: editing} shows the ability of our method in editing tasks.

\subsection{Implementation Details and Benchmark}
\label{subsec:implementation details}
Our method is compatible with ViT models of any size; for simplicity, we adopt ViT-Base as the backbone in our experiments. The model is self-supervised and pre-trained on ImageNet-1K~\cite{deng2009imagenet}. Our implementation mainly follows ViT~\cite{he2022masked}, using the AdamW~\cite{loshchilov2017decoupled} optimiser with a cosine decay learning rate schedule and an initial learning rate of  $1.5\times10^{-4} $. All images are resized and centre-cropped to  $224\times224 $. Full implementation and training details are provided in Supp. Sec. \linkblue{2}.

% Image normalisation is also applied. 
% Since the ViT-base has fewer parameters, it is easier to complete training, meanwhile, it is also more prone to overfitting.

% 有机会加一个所有size的表
% 

\subsection{Performance on the Pre-training Task}
\label{subsec:pre-training}

The pre-training task is formulated as image reconstruction. Figure~\ref{fig:pre-trained} presents qualitative results along with PSNR (Peak Signal-to-Noise Ratio) and SSIM (Structural Similarity Index Measure) scores~\cite{wang2004image}. The results demonstrate that the model can effectively reconstruct images using only isolated and incomplete descriptors.
% fragmentary

\noindent \textbf{Colour segmentation clusters.}
Choosing an appropriate number of clusters is critical for colour segmentation. Too few clusters lead to excessive information loss, degrading performance; too many make the segmented map resemble the original image, oversimplifying the task and limiting the model’s ability to learn meaningful representations. Figure~\ref{fig:num_of_cluster_data} illustrates the visual impact of varying cluster counts and their effect on linear probing performance. The model performs best when the number of clusters is set to $6$.

% When performing image segmentation with K-means, the choice of the number of clusters is crucial. A value too small may not retain sufficient colours, while a value too large can lead to over-segmentation, producing many unnecessary small regions and causing detail leakage. 

% \noindent \textbf{Multi-model Combination.}

% The qualitative results of our reconstruction task are shown in Figure \ref{fig:pretrained task}. The training logging is shown in Appendix A.\ref{fig:training_logger_pretraining}.

% Comparing the reconstructed second features with the source second features, it shows that our model correctly extracts and restores the obtained features. Our model tries to  the scene and object based on the input features. In this case, the provided colour segmentation map (Figure \ref{fig:pretrained colour}) can only provide the dominant hues of the original image. However, from Figure \ref{fig:pretrained construction raw}, we can observe that the generated image reasonably incorporates a variety of colours that are not included in the image segmentation map. We believe this reasoning-like behaviour is related to learning the  representations in the image.

\noindent \textbf{Input independence.}
We evaluate how different inputs independently influence the model's output. Ideally, modifying one input should affect only specific attributes of the output while leaving others largely unchanged. 

We first vary the grey-level histogram, which captures image brightness. We use the SIDD dataset~\cite{abdelhamed2018high}, which offers images under low, normal, and high exposure. We extract all inputs from the normal image. For other exposures, only the histogram is updated, keeping edge and colour descriptors fixed. Figure~\ref{fig:brightness} shows the qualitative results.

To assess the effect of colour segmentation, we use synthetic examples due to the difficulty of sourcing real images with identical edges and brightness but different colours. Starting with an image from the Flickr dataset~\cite{young2014image}, we extract all descriptors. We then alter its colour balance to generate variations and extract new colour segmentation maps, which are combined with the original edge and histogram inputs. Visual results are shown in Figure~\ref{fig:colour}.

% \begin{figure}[t]
%     \centering
%     \begin{subfigure}[t]{0.36\linewidth}
%         \centering
%         \includegraphics[width=\linewidth, keepaspectratio, trim=0 10 0 0, clip]{figures/experiments/brightness.png}
%         \caption{\small Reconstruction result with different illumination brightness level grey-level histogram inputs.}
%         \label{fig:brightness}
%     \end{subfigure}
%     \hfill
%     \begin{subfigure}[t]{0.62\linewidth}
%         \centering
%         \includegraphics[width=\linewidth]{figures/experiments/colour.pdf}
%         \caption{\small Reconstruction with different colour segmentation maps input extracted from the image after different colour balance processing with $f$ is the enhancement factor.}
%         \label{fig:colour}
%     \end{subfigure}
%     \vspace{3mm}
%     \caption{Independence evaluation.}
%     \label{fig:mainfig}
% \end{figure}

% First figure

%  of colour balance processing, black and white image is given if $f=0$, $f=1$ gives the original image.

% \input{figures/experiments/pretained}

\subsection{Representation Analysis on Classification}
\label{subsec:classification}
\smallskip \noindent\textbf{Setting.}
To evaluate the performance of the representations of different models, we conduct classification experiments with fine-tuning and linear probing. In this experiment, we extract representations from the pre-trained encoder and a lightweight header is added for fine-tuning and linear probing. An additional learnable global token is added to the model at the input of the encoder to serve as its state at the output as a global representation. The full implementation detail is shown in Supp. Sec. \linkblue{2}.

To validate the efficacy of our proposed methodology, we conducted comparisons with previous self-supervised methods, Split-brain Autoencoder~\cite{zhang2017split}, MAE~\cite{he2022masked}, and PeCo~\cite{dong2023peco}. To ensure fair comparisons, the backbone of all the models is ViT-base architecture. We replace the Split-brain Autoencoder encoder and decoder with ViT-base architecture, as the original implementation is CNN-based. Meanwhile, we compared the performance with our implementation without pre-training to validate the effectiveness of the pre-training method.

\newcolumntype{M}[1]{>{\centering\arraybackslash}m{#1}}

\newcommand{\textBF}[1]{%
    \pdfliteral direct {2 Tr 0.3 w} %the second factor is the boldness
     #1%
    \pdfliteral direct {0 Tr 0 w}%
}

\newcommand{\down}[1]{\textcolor[RGB]{ 180,12,22  }{\small(↓#1)}}

% \textcolor[RGB]{   180,12,22    |  12,180,22 

\begin{table*}[t]

\begin{minipage}[b]{0.43\linewidth}
\centering
\caption{Classification generalisation in Accuracy (\%).  
%Top-1 class accuracy (\%) reported.
}
\vspace{1mm}
\label{tab:classification}
	\resizebox{\linewidth}{!}{
    \begin{tabular}{l@{\hspace{0.3cm}}|@{\hspace{0.4cm}}c@{\hspace{0.4cm}}c}
     \toprule

  \multirow{2}{*}{Method}&  \multicolumn{2}{c}{Classification (Acc.\%)}  \\
   & Finetuning & Linear Probing \\ 
\midrule
\rowcolor{gray!20} 
scratch, our impl.&    82.0       &          0.1  \\
Split-brain Auto~\cite{zhang2017split}
       &     82.3        &   36.4  \\
MAE~\cite{he2022masked} &     83.6        &   72.4  \\
PeCo~\cite{dong2023peco} &        \textbf{84.5}           &    51.7  \\
\midrule
\NAME~(Ours) &     {83.5}        &     \textbf{74.0}   \\

     \bottomrule
    \end{tabular}
    }
\end{minipage}  % -----------------------------------------
\hspace{0.1cm}
\begin{minipage}[b]{0.55\linewidth}
\centering
\caption{ Transfer learning for classification  and segmentation. 
%  performance on Places dataset
%Top-1 class accuracy (\%) reported.
}
\vspace{1mm}
\label{tab:image classification transfer learning}
	\resizebox{\linewidth}{!}{
    \begin{tabular}{l@{\hspace{0.3cm}}|@{\hspace{0.3cm}}c@{\hspace{0.3cm}}c}
     \toprule
% M{2cm}

  \multirow{2}{*}{Method}& Classification (Acc.\%) &   Segmentation (mIoU)  \\
   & Places dataset &  ADE20K dataset \\
\midrule
\rowcolor{gray!20} 
scratch, our impl.  & 79.9 & 47.7 \\
Split-brain Auto~\cite{zhang2017split}
        & 81.4  & 44.4\\
MAE~\cite{he2022masked}  & 82.5  & 48.4\\
PeCo~\cite{dong2023peco}  & 82.4 & 48.7\\
\midrule
\NAME~(Ours)      & \textbf{82.7} & \textbf{49.7}\\

     \bottomrule
    \end{tabular}
    }

\end{minipage}
% \vspace{-0.3cm}
\end{table*}

\smallskip \noindent\textbf{Results.}
In Table \ref{tab:classification}, we compare the quantitative results of different methods of fine-tuning and linear probing.
For fine-tuning, the performance differences among all methods are not significant; our method is comparable to MAE~\cite{he2022masked} and slightly inferior to PeCo. However, our model performs better than the other models in linear probing, which indicates that our model can extract useful underlying representations from images better.

\subsection{Representation Analysis on Transfer Learning}
\label{subsec:transfer learning}
% 
% We focus on evaluating the classification performance of features extracted by our model. To assess this, we freeze the feature extraction encoders and the fusion module, and append a linear logistic regression classifier to the output features. We trained the linear layer and validated it on ImageNet~\cite{imagenet_cvpr09}, Places~\cite{zhou2017places}, and PASCAL~\cite{everingham2010pascal}.
To further evaluate the performance of the extracted representation, we used the classification task pre-trained encoder from Section \ref{subsec:classification} to assess transfer learning in downstream tasks.

First, we tested its classification generalisation on other datasets. Additionally,  we conducted a Semantic Segmentation experiment on ADE20K~\cite{zhou2019semantic}, leveraging the segmentation head from Segformer~\cite{xie2021segformer}, which is a lightweight decoder composed solely of MLPs.     The detailed structure is provided in the supplementary material.

Table~\ref{tab:image classification transfer learning} presents transfer learning results on classification and segmentation on Places~\cite{zhou2017places} and ADE20K~\cite{zhou2019semantic} datasets. 
Our model achieves the best transfer performance, \eg~49.7 mIoU in semantic segmentation, outperforming MAE and PeCo by 1.3 and 1.0, respectively. This advantage may arise from pre-training on visual descriptors rather than raw images, encouraging the model to learn underlying structures instead of dataset-specific biases. 

% Compared to existing self-supervised pre-training models, o
% The transfer learning performance for  segmentation are reported in Figure \ref{tab:semantic segmentation transfer learning}. 

\subsection{Representation for Visual Restoration}

\label{subsec: generation}
\smallskip
\noindent\textbf{Setting.} 
In this section, we utilise image restoration tasks to intuitively demonstrate the effectiveness of our learned representations. In this task, we combine with pre-trained Stable Diffusion 1.5~\cite{rombach2022high}. Our learnt representation encompasses both global and local representations. We employ the global representation combined with the original text embedding, inspired by IP-Adapter~\cite{ye2023ip-adapter}, providing an overall condition for the generated image. Meanwhile, the local representations are integrated via ControlNet~\cite{zhang2023adding} to guide the structural generation within the UNet model, enabling precise control in the image regions.

To enable an effective comparison with existing guidance methods, ControlNet~\cite{zhang2023adding}, T2I-Adapter~\cite{mou2024t2i}, ControlNet++~\cite{li2025controlnet}, we use the edge map and segmented colour map directly as inputs for these baseline methods. We also input the grey-level histogram directly through IP-Adapter. However, we found that using only the grey-level histogram in IP-Adapter might limit the model ability during generation. For further comparison, we additionally employ BLIP~\cite{li2022blip} to generate a text prompt, which is then fed into the text encoder.  %Results are shown in Figure \ref{fig:generation}.
% \begin{table}
% \centering
% \caption{Quantitative results on image restoration. The best results are highlighted in \textbf{bold}.}
% \label{tab:restoration}
% \vspace{2mm}
% \renewcommand{\arraystretch}{1.0} % Increase row height for better readability
% \setlength{\tabcolsep}{8pt} % Adjust column spacing
%  \resizebox{0.48\textwidth}{!}{
% \begin{tabular}{l@{\hspace{0.4cm}}|@{\hspace{0.4cm}}c@{\hspace{0.4cm}}|@{\hspace{0.4cm}}c@{\hspace{0.4cm}}|@{\hspace{0.4cm}}c}
% \toprule

%   Method&  Prompt & PSNR & SSIM \\ 
% \midrule
% \multirow{2}{*}{ControlNet~\cite{zhang2023adding}} & w/o prompt&        17.34& 0.6374   \\
%  & w/ prompt&        16.52& 0.6051   \\
%  \midrule
% \multirow{2}{*}{T2I Adapter~\cite{mou2024t2i}}   &  w/o prompt &        17.69& 0.5421   \\
%  & w/ prompt&        17.30& 0.5459   \\
%   \midrule
%    \multirow{2}{*}{ControlNet++~\cite{li2025controlnet}} & w/o prompt&        19.94& 0.6549   \\
% &  w/ prompt&        19.50& 0.6399   \\

% \midrule
% \NAME~(Ours) & w/o prompt   &       \textbf{26.56} & \textbf{0.8664}   \\
%      \bottomrule
%     \end{tabular}
% }
% \end{table}

\begin{wraptable}{r}{0.4\textwidth}
\centering
\vspace{-4mm} % Adjust vertical spacing as needed
\caption{ Image restoration results. }  % Quantitative  on
\label{tab:restoration}
\vspace{2mm}
\renewcommand{\arraystretch}{1.0}
\setlength{\tabcolsep}{6pt}
% \resizebox{0.4\textwidth}{!}{
\resizebox{0.4\textwidth}{!}{
\begin{tabular}{lccc}
\toprule
Method & Prompt & PSNR & SSIM \\
\midrule
\multirow{2}{*}{ControlNet~\cite{zhang2023adding}} & w/o & 17.34 & 0.6374 \\
 & w/ & 16.52 & 0.6051 \\
% \midrule
\multirow{2}{*}{T2I Adapter~\cite{mou2024t2i}} & w/o & 17.69 & 0.5421 \\
 & w/ & 17.30 & 0.5459 \\
% \midrule
\multirow{2}{*}{ControlNet++~\cite{li2025controlnet}} & w/o & 19.94 & 0.6549 \\
 & w/ & 19.50 & 0.6399 \\
\midrule
\NAME~(Ours) & w/o & \textbf{26.56} & \textbf{0.8664} \\
\bottomrule
\end{tabular}
}
\vspace{-2mm}
\end{wraptable}

\begin{figure*}[t]
    \centering
    \vspace{-0.05cm}
    \includegraphics[width=\textwidth, keepaspectratio]{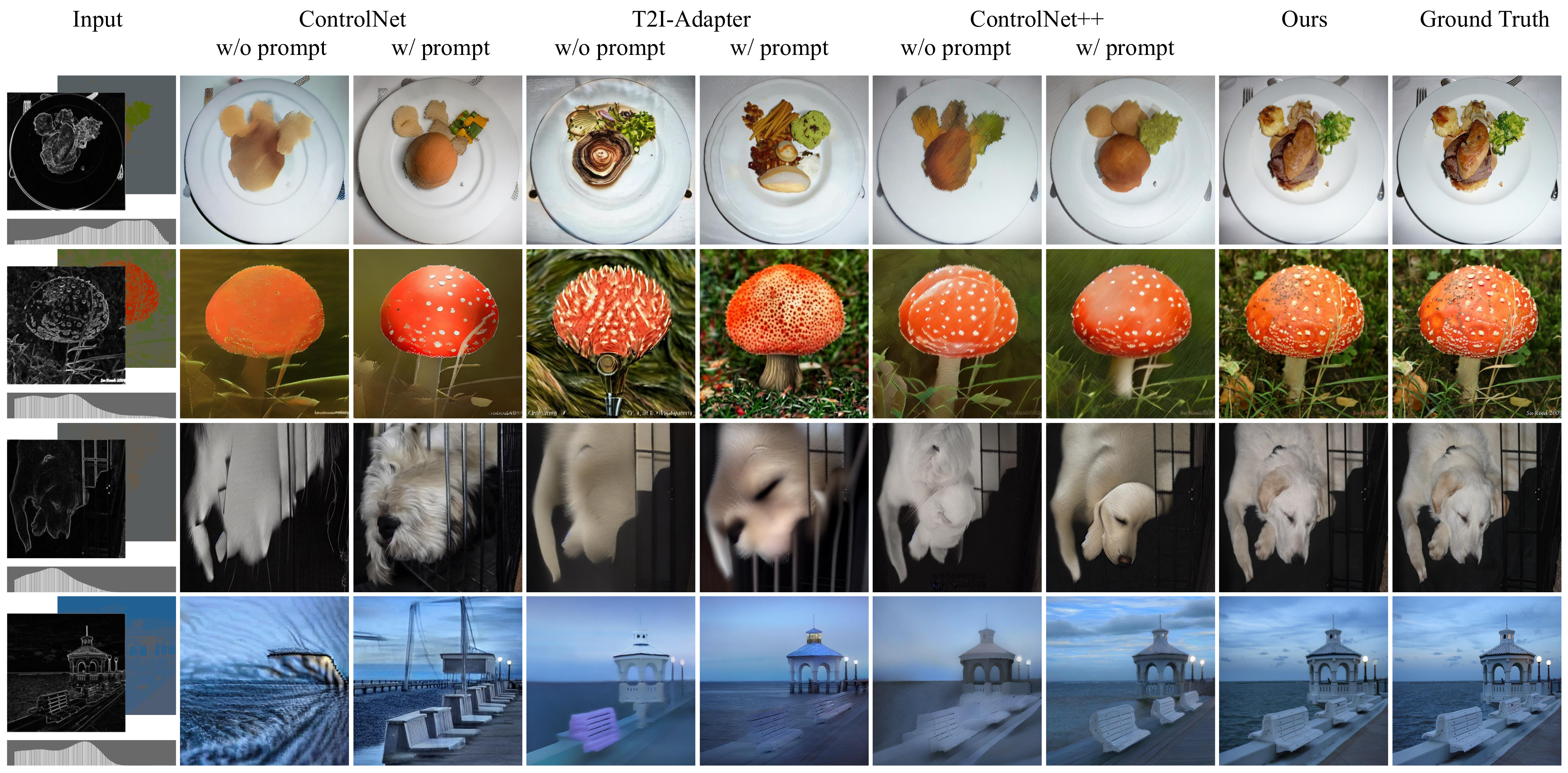}
    \vspace{1mm}
    \caption{\textbf{Comparison of image restoration results} using Stable Diffusion 1.5 guided by ControlNet~\cite{zhang2023adding}, T2I-Adapter~\cite{mou2024t2i}, ControlNet++~\cite{li2025controlnet}, and our method, with the ground truth shown in the final column. The condition inputs are edge, segmented colour map, and grey-level histogram inputs, shown in the first column. Each method, except for our method, displays results with and without prompts, where prompts are generated by BLIP~\cite{li2022blip}. }
    \label{fig:generation}
    % \vspace{-0.4cm}
\end{figure*}

\smallskip \noindent\textbf{Results.}
As shown in Table~\ref{tab:restoration} and Figure~\ref{fig:generation}, our method reconstructs high-quality images, preserving both global features and local details better than baseline approaches. This highlights the advantage of our structured representations over raw inputs, improving model interpretability and generalizability. Moreover, our approach effectively manages multiple representation controls, overcoming the limitations of traditional methods. 

\vspace{-3mm}
\subsection{Representation for Visual Editing}
\label{subsec: editing}
Our approach benefits from separating visual descriptors in advance. In image editing tasks, it allows us to edit these descriptors with much greater ease than directly manipulating and controlling the image. In this section, we demonstrate how simple adjustments to the segmented colour map and grey-level histogram enable controllable image editing without any additional training or modification on models.
% \smallskip \noindent\textbf{Modification via Segmented Colour Map.}
We followed the setting of Section \ref{subsec: generation}, and only edited the original segmented colour map to edit the colour regions of the generated image and keep the other inputs the same. The region of the colour map will be recoloured only if it would otherwise conflict with the edge.

\begin{figure*}[t]
    \centering
        \vspace{-0.05cm}
    \includegraphics[width=\textwidth, keepaspectratio]{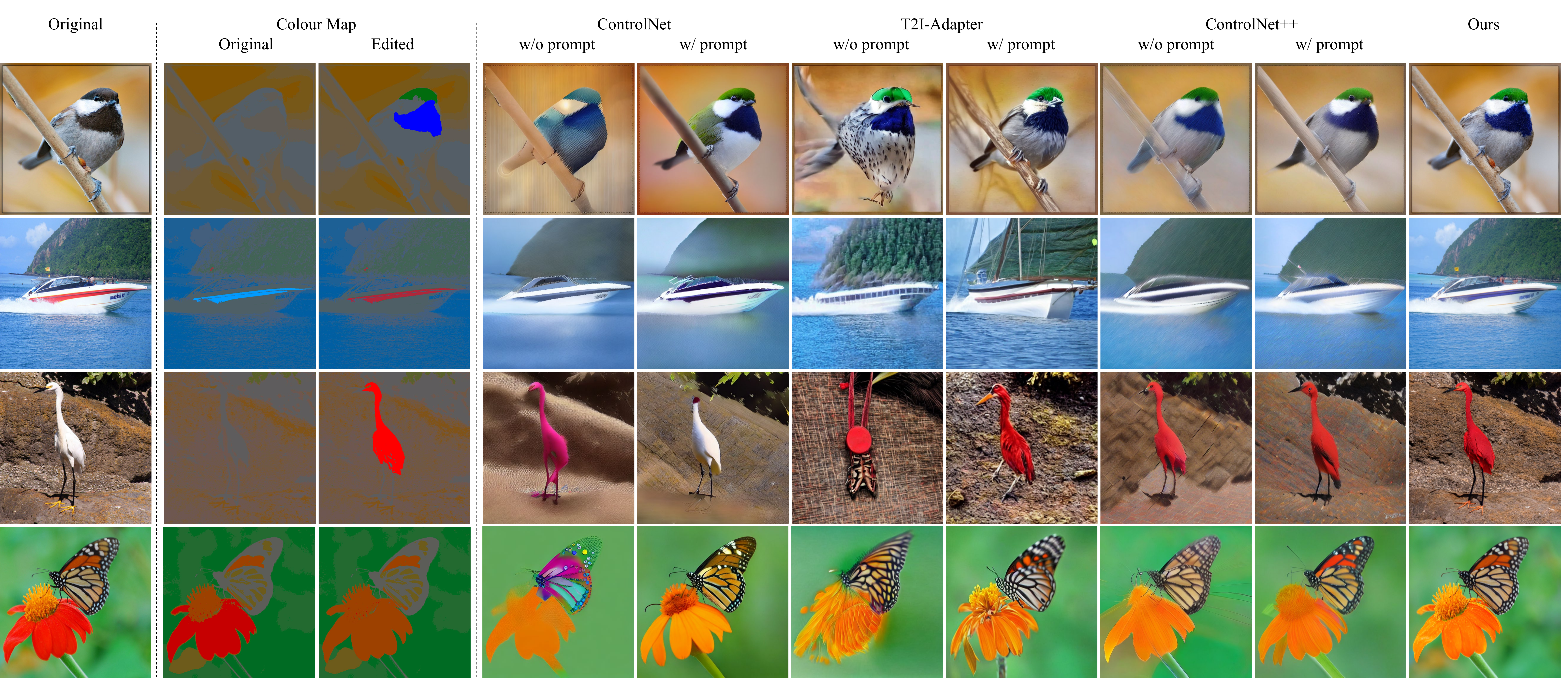}
    \vspace{0.5mm}
    \caption{\textbf{Comparison of colour editing results} using modified segmented colour maps as input to Stable Diffusion 1.5, guided by ControlNet~\cite{zhang2023adding}, T2I-Adapter~\cite{mou2024t2i}, ControlNet++~\cite{li2025controlnet}, and our method. The first column shows original images, followed by original and edited colour maps. For each baseline, results are shown with and without prompts generated by BLIP using edited colour keywords. Our method requires \textbf{no prompts}.}
    \vspace{-3mm}
    \label{fig:editing result}
        % \vspace{-0.2cm}
\end{figure*}

As illustrated in Figure \ref{fig:editing result}, modifying the segmented colour map provides a straightforward way of directing the model to edit the original image with precise control. Notably, our method maintains image harmony while performing controlled editing. We also have modified the grey-level histogram for image editing to adjust the brightness of the output, which is shown in Supp. Sec. \linkblue{5.2}.
We further conducted a survey based on human perception for the qualitative analysis of image editing tasks in Supp. Sec. \linkblue{5.1}. The result shows that our method significantly outperforms the others in all attributes.

% \smallskip \noindent\textbf{Modification via Gray-Level Histogram.}
% \input{figures/experiments/brightness_result}
% We also explored image brightness editing through adjustments to the gray-level histogram while maintaining the same setup as before but adjusting gray-level histogram. We adjust gray-level histogram by adjusting the exposure compensation with following equation:
% \begin{equation}
%     i' = i \times 2^\epsilon,
% \end{equation}
% where i is the l channel in the image. We also did histogram equalization for the original image with equation:
% \begin{equation}
%     i' = \text{round}\left((L - 1) \cdot \frac{\sum_{j=0}^{i} n_j}{N}\right),
% \end{equation}
% where L is the number of gray level. Because signal leakage in Stable Diffusion 1.5 training makes it only be able to generate images with medium brightness~\cite{lin2024common}, we used a variant with v\_prediction and enable a zero signal-to-noise ratio.

% As shown in Figure \ref{fig:brightness generation}, the edited result shows the similar effect like classic image enhancement methods while preserving the coherence of other visual elements. The generated repplt follows the input gray-level histogram.
\vspace{-3mm}
\section{Discussion}
\vspace{-3mm}
\label{sec:discussion}

% In this section, we analyse the results presented in Section \ref{sec:experiments}. Our pre-training method, feeding more abstract and multi-modal descriptors instead of images, encourages the model to learn the underlying representation better by this reasoning-like prediction. The results in both high-level and low-level tasks validate the effectiveness of our method.

\smallskip \noindent\textbf{Mask-free training.}
Unlike random pixel dropout, our mask-free approach eliminates the need for masking ratio tuning.   From an information-theoretic perspective, it leverages only 17\% of the RGB signal, less than Split-Brain Autoencoder (50\%), MAE (25\%), and PeCo (60\%), which encourages the model to learn better representations~\cite{bao2021beit}. See Supp. Sec. \linkblue{3}.
As shown in Section~\ref{subsec:classification}, our method slightly underperforms baselines after fine-tuning but notably outperforms them under linear probing, indicating that our method learns more explicit representations. It also generalises across datasets and tasks (Section~\ref{subsec:transfer learning}), while in generation tasks (Section~\ref{subsec: generation}). The learnt representation adapts seamlessly to other models.

% classical descriptors naturally emphasize meaningful global and local structures.

% This sparse input encourages the network to capture core visual concepts, leading to stronger representations.

% \input{figures/experiments/failure_case}

\smallskip \noindent\textbf{Decouple controllable attributes.}
Section~\ref{sec:experiments} demonstrates \ourmodel's ability to isolate and control specific attributes (\eg geometry, colour, and illumination) via decoupled visual inputs. In Section~\ref{subsec:pre-training}, we show that these inputs remain independent of the generated outputs during pre-training. Section~\ref{subsec: editing} further illustrates that our descriptors can independently and effectively guide the generation process without requiring any model modification, enabling precise control over individual attributes without altering the whole.

% \smallskip \noindent\textbf{Limitations.}
% While \ourmodel~produces promising results, certain areas require refinement. Although our method achieves notable decoupling, its reliance on classical descriptors may limit adaptability to highly complex images with fine details, primarily due to the limitations of segmented colour representation. Image segmentation inherently results in the loss of subtle colour variations and small-region details, reducing the model sensitivity to these areas, which may impact performance on specific tasks. Figure~\ref{fig:failure cases} illustrates failure cases in the restoration task, where the red spots on the newt (left) and the spots on the butterfly (right) are partially missed.

% \input{sections/further}

\begin{figure}[t]
\begin{adjustbox}{center,raise=-2cm}
\resizebox{0.95\columnwidth}{!}{
    \begin{tabular}{c@{\hspace{0.5em}}c@{\hspace{0.5em}}c@{\hspace{0.5em}}c}
         \LARGE Ground Truth & \LARGE Ours & \LARGE  Ground Truth & \LARGE  Ours \\
        \includegraphics[width=0.5\linewidth, keepaspectratio]{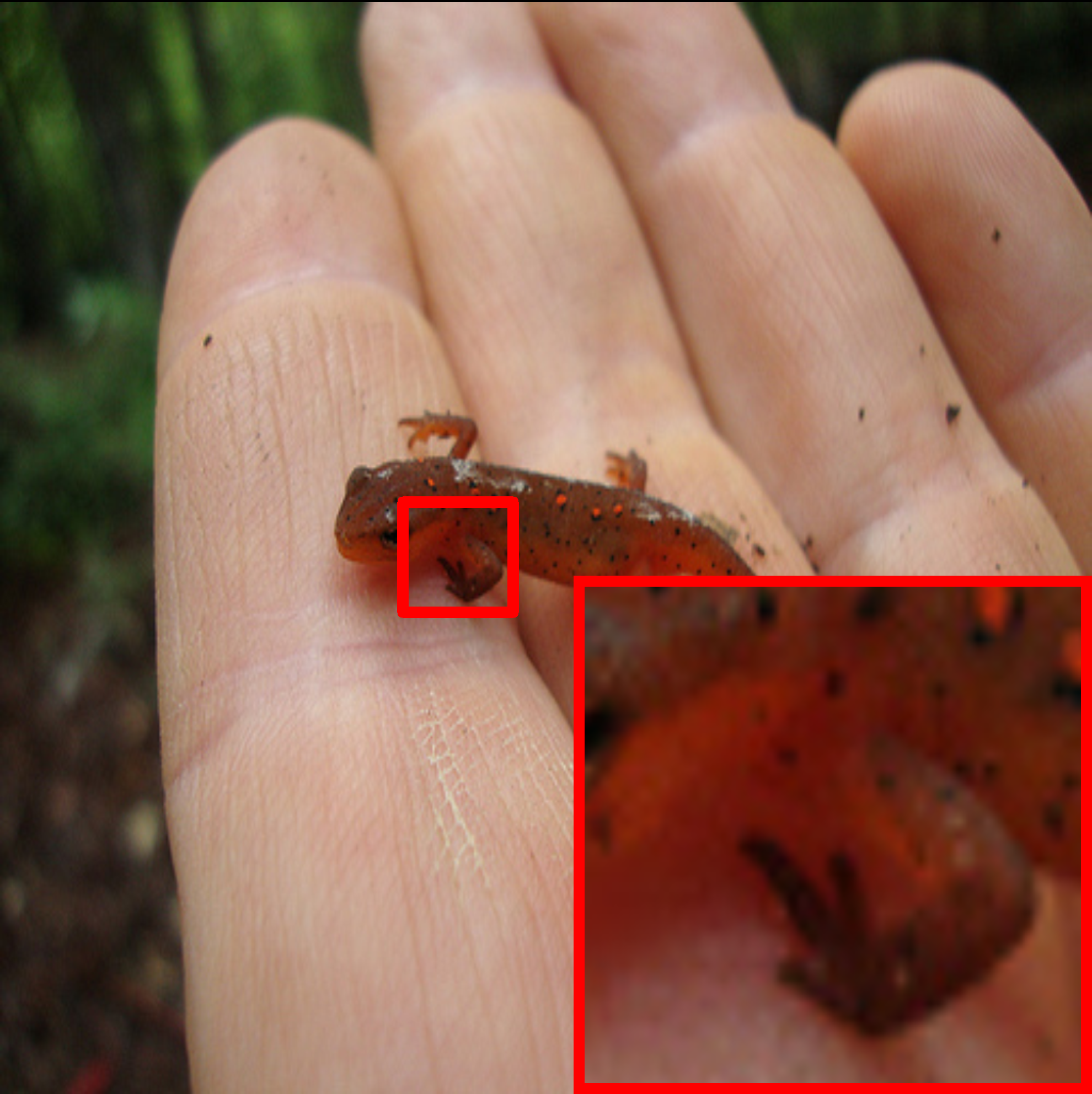} &
        \includegraphics[width=0.5\linewidth, keepaspectratio]{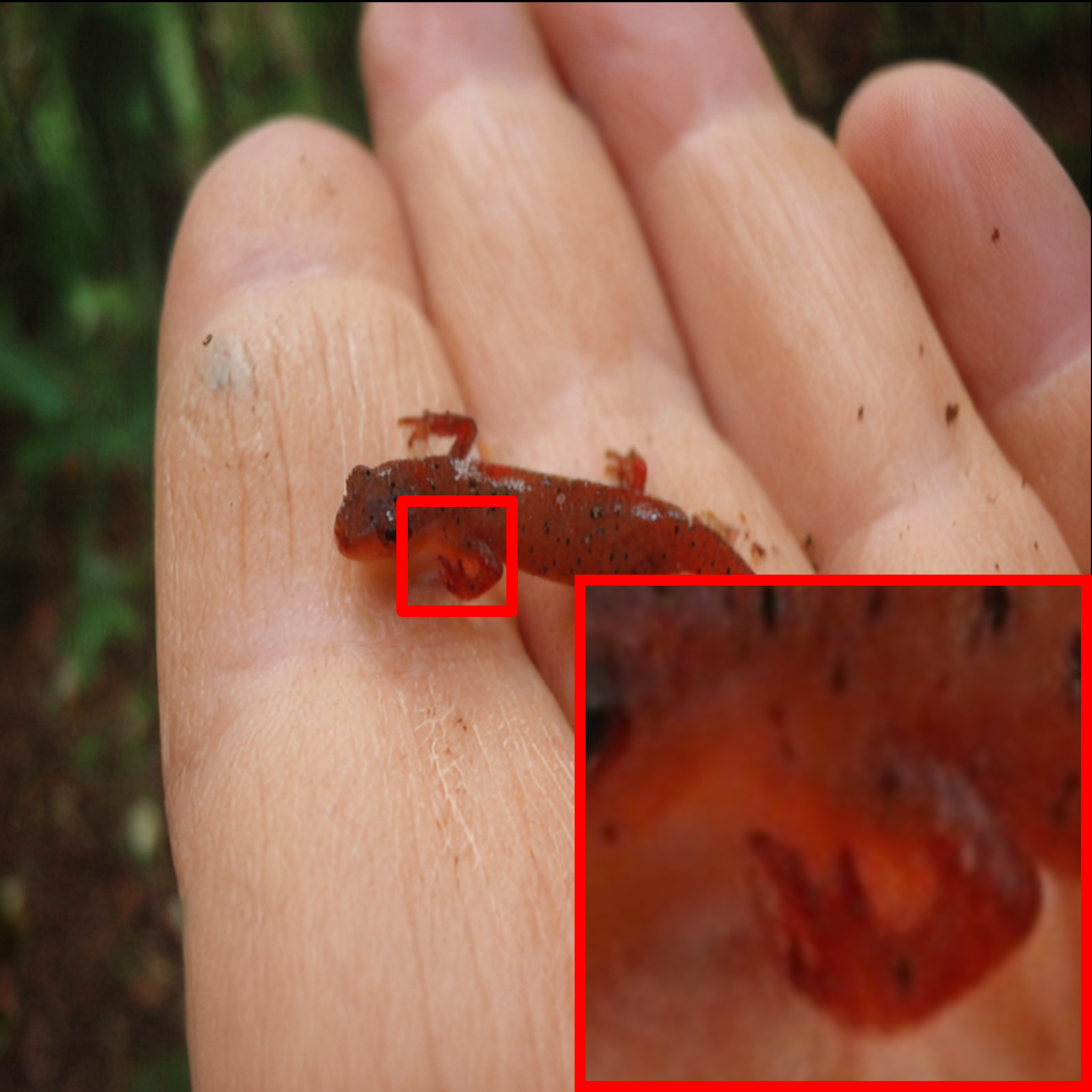} &
        \includegraphics[width=0.5\linewidth, keepaspectratio]{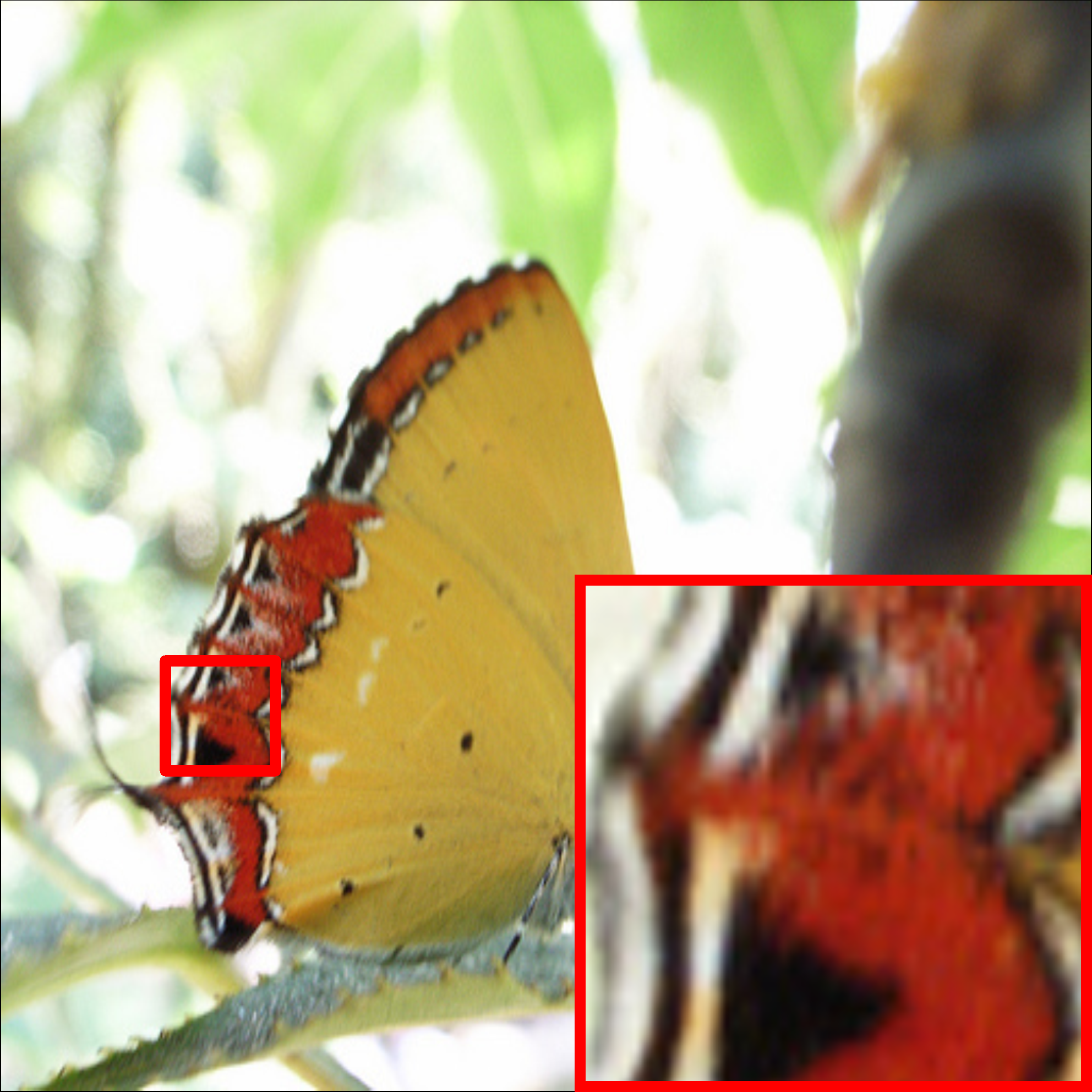} &
        \includegraphics[width=0.5\linewidth, keepaspectratio]{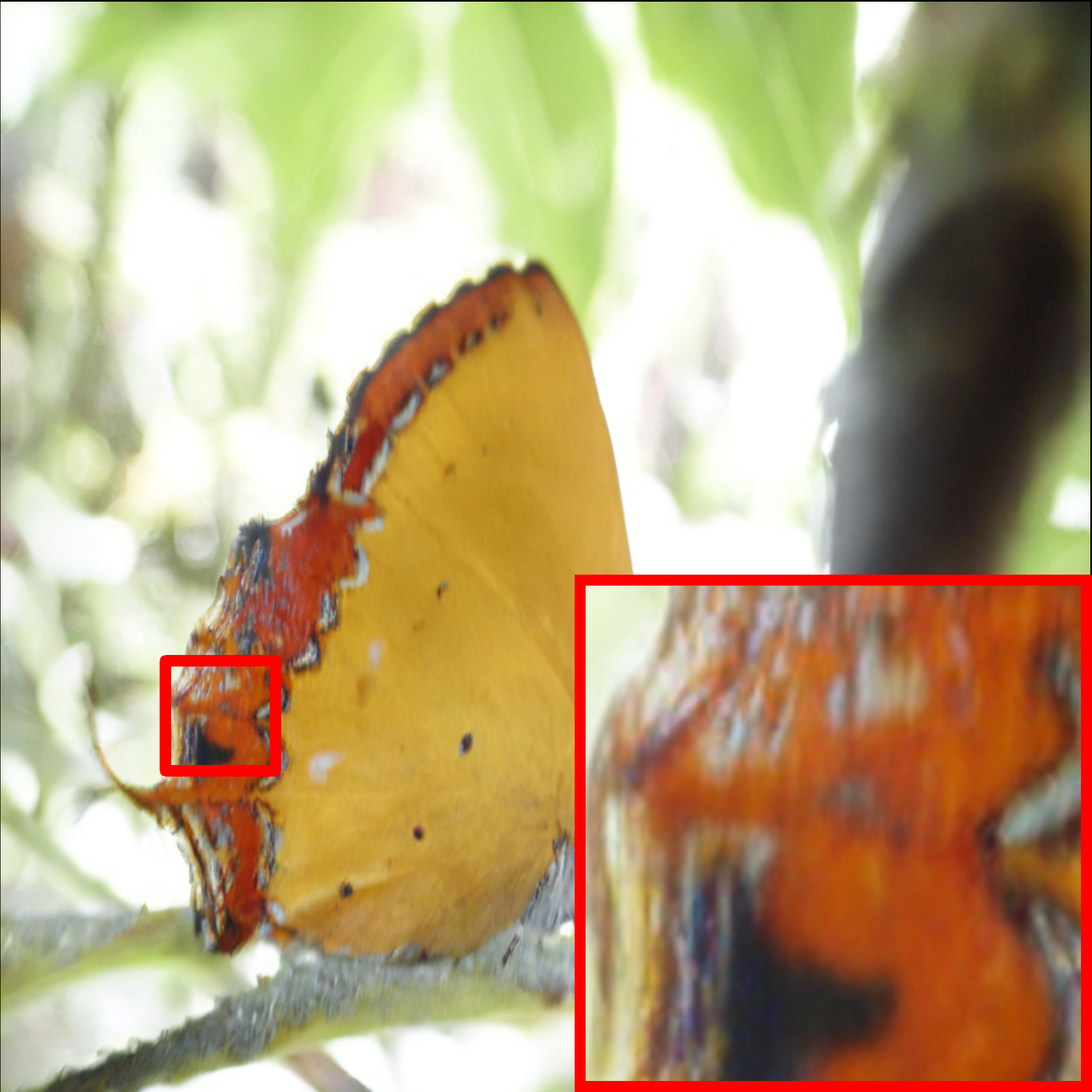} 
    \end{tabular}
}
\end{adjustbox}
\vspace{1.5mm}
     \caption{Failure cases on the visual restoration task.}
\vspace{-4.5mm}
\label{fig:failure cases}
\end{figure}

% \begin{figure}
%     \centering
%     \includegraphics[width=\linewidth, keepaspectratio]{figures/experiments/failure case.pdf}
%     \vspace{-0.3cm}
%     \caption{Failure cases on the restoration task.}
%     \label{fig:failure cases}
% \end{figure}

\section{Limitations and Future Work}
\label{sec:limitation}

This work advances the understanding of disentangled visual representation learning, but the choice of visual descriptors warrants further research. Segmented colour maps can miss small-area chromatic details, reducing sensitivity to subtle regions; Figure~\ref{fig:failure cases} shows restoration failures where the spots on the newt (left) and the wing markings on the butterfly (right) are partially missed. In addition, the grey-level histogram discards spatial layout. While this design choice was intended to challenge the model to learn essential image information without direct access to the original visuals, it proved overly difficult. As shown in Figure~\ref{fig:illustration}, the human-drawn depiction of a dog incorrectly renders areas expected to be white as grey. 

Task-specific descriptors may be more suitable for certain datasets. For portrait datasets, human keypoints may be more suitable to describe them, which may also be easier to edit by dragging the points. In this paper, we focus more on the fundamental feasibility of this approach and therefore do not discuss specific applications in detail.

% Another significant challenge encountered during this research pertains to the multi-task training framework. While we have proposed some solutions in Section \ref{appsubsec:Loss Weighting}, achieving optimal performance across multiple tasks simultaneously is still complex. Moreover, this balancing action can still sometimes lead to compromised performance in certain tasks, as adjustments made to benefit one task may inadvertently impact the effectiveness of the model on another.

\section{Conclusion}
In this work, we investigated the challenge of decoupling visual content by leveraging classical visual descriptors within learning-based paradigms, which have been largely overlooked in deep models. Inspired by the human ability to infer and reconstruct visual content from incomplete descriptor combinations, we explore whether deep models can emulate this capacity and propose \ourmodel, a new visual decoupling paradigm. 
Our experiments demonstrate that the learned representations not only excel in low-level tasks such as classification and segmentation but also effectively guide high-quality image generation and editing through decoupled descriptor inputs. Furthermore, this approach opens new possibilities for advanced image manipulation and encourages further research into decoupling-based models that integrate conventional visual descriptors. 
We hope this work provides a fresh perspective on visual decoupling and inspires future research on its extensive applications across various visual tasks.

\section*{Acknowledgements}

This project is partially supported by the Royal Society grants (SIF\textbackslash R1\textbackslash231009, IES\textbackslash R3\textbackslash223050) and an Amazon Research Award.
The computations in this research were performed using the Baskerville Tier 2 HPC service. Baskerville was funded by the EPSRC and UKRI through the World Class Labs scheme (EP\textbackslash T022221\textbackslash1) and the Digital Research Infrastructure programme (EP\textbackslash W032244\textbackslash1) and is operated by Advanced Research Computing at the University of Birmingham.

% we designed a \ourmodel that feeds representative but incomplete visual properties into the model, and asked the model to reconstruct the visual information. 

% Unlike conventional methods that use pixel-level reconstruction loss as a supervision signal, we chose to use the same incomplete visual properties but extracted from the reconstructed images. Interestingly, experimental analysis showed the effectiveness of such an approach in learning  representations. The learned  representations also behaved differently on different downstream vision tasks, suggesting the applicability of the learned representations and the potential of the proposed  visual representation learning approach.

\bibliography{egbib}
\end{document}